\documentclass[lettersize,journal]{IEEEtran}
\usepackage{amsmath,amsfonts}
\usepackage{hyperref}
\hypersetup{hidelinks,
	colorlinks=true,
	allcolors=black,
	pdfstartview=Fit,
	breaklinks=true}
\usepackage{algorithmic}
\usepackage{array}
\usepackage[caption=false,font=normalsize,labelfont=sf,textfont=sf]{subfig}
\usepackage{textcomp}
\usepackage{stfloats}
\usepackage{url}
\usepackage{multirow}

\usepackage{verbatim}
\usepackage{graphicx}
\usepackage{graphicx,times,amsmath} 
\usepackage{graphicx}
\usepackage{amsmath,mathrsfs}
\usepackage{amssymb}
\usepackage{booktabs}
\usepackage{stfloats}
 \usepackage{cite}
 
\usepackage{color}
\usepackage{longtable}
\definecolor{myorange}{RGB}{255,165,0}
\newcommand{\orange}[1]{\textcolor{myorange}{#1}}

\newcommand{\red}[1]{\textcolor{red}{#1}}
\usepackage[linesnumbered,ruled]{algorithm2e}
\usepackage{bm}
\usepackage{wrapfig}
\usepackage{threeparttable}
\usepackage{pifont}
\usepackage{utfsym}

\usepackage{enumitem}
\setenumerate[1]{itemsep=0pt,partopsep=0pt,parsep=\parskip,topsep=5pt}
\setitemize[1]{itemsep=0pt,partopsep=0pt,parsep=\parskip,topsep=5pt}
\setdescription{
    itemsep=0pt,partopsep=0pt,parsep=\parskip,topsep=5pt}

\begin{document}

\title{RoboGPT: an LLM-based Embodied Long-term Decision Making agent for Instruction Following Tasks}


\author{ Yaran Chen, Wenbo Cui, Yuanwen Chen, Mining Tan, Xinyao Zhang, Jinrui Liu,  Dongbin Zhao, He Wang, and Haoran Li
\thanks{This work has been submitted to the IEEE for possible publication.
Copyright may be transferred without notice, after which this version
may no longer be accessible.}
\thanks{
Y. Chen, W. Cui, M. Tan, X. Yao, J. Liu, D. Zhao and Haoran Li are with the State Key Laboratory of Multimodal Artificial Intelligence Systems,
Institute of Automation, Chinese Academy of Sciences, Beijing, China (dongbin.zhao@ia.ac.cn)}
\thanks{H. Wang is with Center on Frontiers of Computing Studies, Peking University }
}

\markboth{Journal of IEEE Transactions on Cognitive and Developmental Systems, Vol. 00, No. 0, Month 2024}
{First A. Author \MakeLowercase{\textit{et al.}}: Bare Demo of IEEEtran.cls for IEEE Journals of IEEE Transactions on Cognitive and Developmental Systems}
\maketitle

\begin{abstract}

Robotic agents are tasked with mastering common sense and making long-term sequential decisions to execute daily tasks based on natural language instructions.  Recent advancements in Large Language Models (LLMs) have catalyzed efforts for complex robotic planning. However, despite their superior generalization and comprehension capabilities, LLM task plans sometimes suffer from issues of accuracy and feasibility. 
To address these challenges, we propose RoboGPT,
an agent specifically designed to make embodied long-term decisions for instruction following tasks. RoboGPT integrates three key modules:
1) RoboPlanner, an LLM-based planning module equipped with 67K embodied planning data, breaks down tasks into logical subgoals. We compile a new robotic dataset using a template feedback-based self-instruction method to fine-tune the Llama model. RoboPlanner with strong generalization can plan hundreds of instruction following tasks;
2) RoboSkill, customized for each subgoal to improve navigation and manipulation capabilities; 
3) Re-Plan, a module that dynamically adjusts the subgoals based on real-time environmental feedback. By utilizing the precise semantic map generated by RoboSkill, the target objects can be replaced by calculating the similarity between subgoals and the objects present in the environment.
Experimental results demonstrate that RoboGPT exceeds the performance of other state-of-the-art (SOTA) methods, particularly LLM-based methods, in terms of task planning rationality for hundreds of unseen daily tasks and even tasks from other domains.
 \end{abstract}




\section{Introduction}

Embodied AI tasks, encompassing visual navigation and robotic manipulation, have progressed rapidly \cite{RT-2, LU2021140, li2024stabilizing}. Anticipated future robots are projected to assist humans in executing complex daily tasks by following natural language instructions, such as `\textit{make dinner}' or `\textit{wash dishes}' \cite{RT-2,RT1brohan2022}. Existing methodologies, including template planning \cite{prompterinoue2022,Filmmin2021} and expert-guided planning \cite{ET-PashevichS021,MCR-agent2023}, have demonstrated some success in handling seven types of instruction following tasks. 
Nevertheless, current agents fall short in fully comprehending instruction tasks, including aspects like object quantities, prefix content, and object dependencies. 

\begin{figure*}[t!]
   \centerline{
      \includegraphics[width=0.9\linewidth]{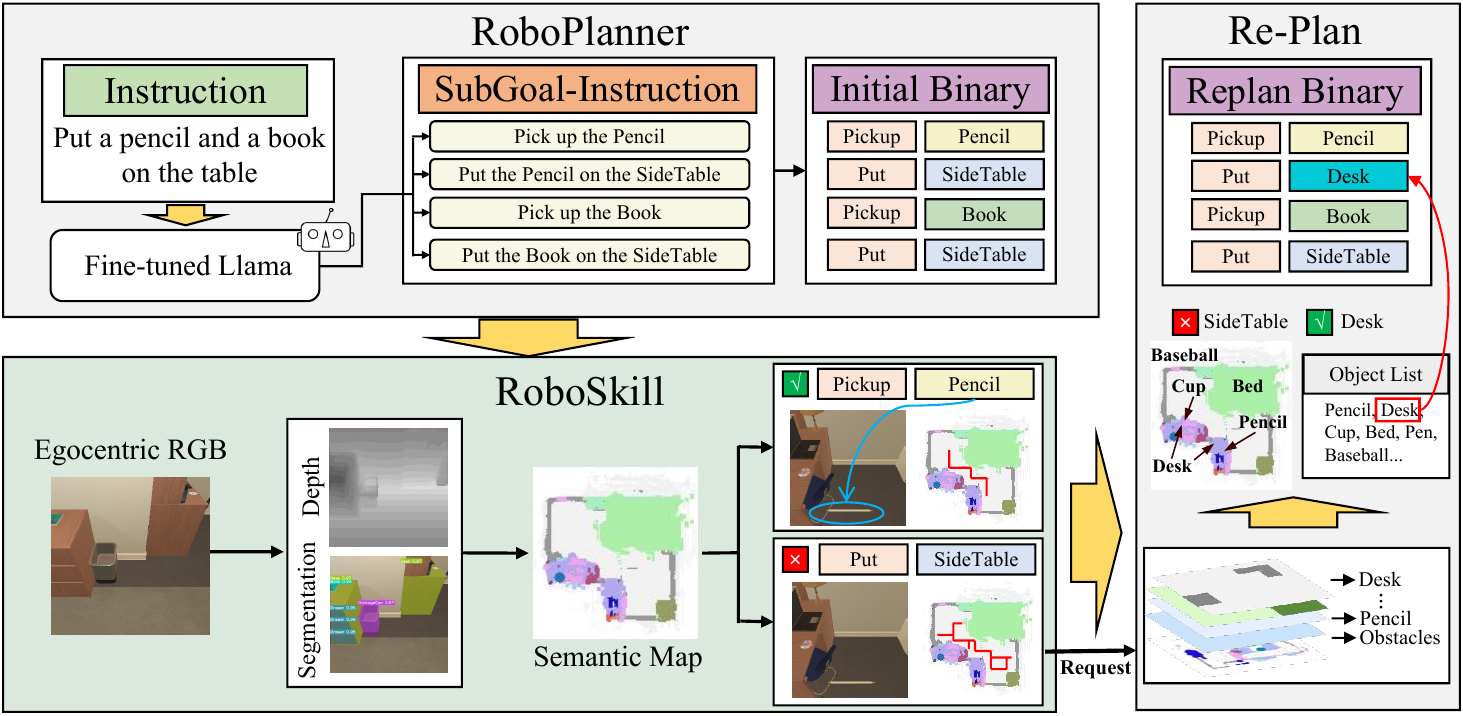}
   }
\caption{The architecture of RoboGPT. RoboPlanner decomposes an instruction into logical subgoals. RoboSkill encompasses navigation, manipulation, and interaction with the environment sequentially based on subgoals. If a subgoal fails, Re-Plan receives feedback and generates a new plan based on environmental information, e.g., replace `\textit{SideTable}' with `\textit{Desk}'.}
   \label{fig:architecture}
\end{figure*}

Large Language Models (LLMs) have made significant progress in the field of natural language processing \cite{touvron2023llama}. Due to their extensive internalized world information, LLMs can solve complex embodied planning problems \cite{llm-plannersong2022} more generically than template-based methods \cite{prompterinoue2022}. However, generic LLMs are overly broad and lack robotics expertise, resulting in plans that are frequently unfeasible for direct implementation by robots \cite{saycanahn2022can,lin2023text2motion,llm-plannersong2022}.  
For instance, given the task `\textit{If you are a robot and give me a plan of: slice an apple}', the generated plan is `\textit{gather material', `wash your hands', `prepare the apple', `...}', which is not executable for a robot. Even LLM-Planner with a designated prompt \cite{llm-plannersong2022} produces some illogical planning of complex daily tasks. 

To address the accuracy issue of LLM planning, this paper seeks to enhance the planning capabilities of LLMs by integrating expertise from the robotics domain. We aim to fine-tune and enhance the LLM planning process to ensure both logical validity and optimal execution. To overcome the lack of domain-specific data, we create a specialized robotic dataset consisting of 67K embodied commands covering complex robotic activities. Data collection and fine-tuning are key steps in this enhancement process. We employ an automatic approach, utilizing a self-instruction method that enables ChatGPT to generate a batch of data. However, we find that over 60\% of the generated data exhibits issues, predominantly incorrect logical relationships in the planning. To address this, we introduce a template feedback mechanism that prompts ChatGPT to introspect and rectify the generated data. The model initially identifies the data type and then inputs the relevant considerations for that type. This process guides ChatGPT to autonomously correct erroneous planning, substantially reducing the labor associated with manual data correction. Unlike LLM-Planner, RoboGPT trained on the above data: 1) can understand the prefix content to modify the planning subgoals according to the environments; 2) can understand the object dependencies to find invisible objects that are in containers, e.g., `\textit{Put an apple from the microwave into the garbage}', RoboGPT plans `\textit{find the microwave first}', while current methods directly find apples, failing to find them always; 3) can understand object quantities to handle tasks with more than two objects beyond other methods. 

The implementation of the template feedback mechanism not only guarantees the production of high-quality data but also promotes the formation of logical thought chains, greatly improving the accuracy of the generated data. The fine-tuned RoboPlanner, trained on this refined dataset, demonstrates robust generalization capabilities and outperforms other planning methods.
Ultimately, the LLM fine-tuned using this dataset exhibits improved performance in embodied instruction following tasks. This innovative approach to data collection and refinement combines with the unique application of a template feedback mechanism and constitutes a significant contribution to our work.

During instruction following tasks, the key to addressing feasibility issue is to enable the agent to perceive and consider the environment.
Mapping instruction targets to objects in the environment remains a challenge, i.e., the instruction nomenclature diversity challenge. For example, the task is `\textit{Put a pencil and a book on the table}', while common table types in the environment are `\textit{sidetable}', `\textit{diningtable}', and `\textit{desk}'. The previous methods \cite{prompterinoue2022,Filmmin2021} typically predict a type of table `\textit{sidetable}' based on experience and plan `\textit{find sidetable}'. However, when the `\textit{sidetable}' is unavailable in the environment, the agent fails to locate a suitable table, leading to task failure. To address this problem, we propose introducing environment feedback and re-planning to align the environment objects with instructed task.

In summary, our primary contributions are as follows:

(1) To address LLM planning's accuracy issue, we create an embodied instruction planning dataset and propose RoboPlanner. This planner demonstrates robust generalization capabilities, enabling it to plan for hundreds of daily tasks. We also design a template-feedback mechanism, enabling the LLM to autonomously generate and modify planning data, thereby significantly reducing human effort in data correction. This is, to the best of our knowledge, the first high-quality, large-scale dataset for daily instruction planning task in the field of robotics.

(2) Re-Plan is developed to address the feasibility issue of LLM planning. It can dynamically adapt to the environment, effectively tackling the challenges posed by nomenclature diversity in instruction following tasks. Moreover,  Re-Plan utilizes more accurate environmental information from the integrated segmentation model, thereby enhancing success rate (SR) in ALFRED tasks.

(3) We develop RoboGPT designed for handling complex daily tasks, including RoboPlanner, RoboSkill, and Re-Plan. This agent exhibits superior performance compared to state-of-the-art (SOTA) methods on both the ALFRED benchmark (SR: 60.7\% versus 50.3\%) and tasks requiring generalization (SR: 78.0\% versus 34.0\%).

\section{Related Work}

\textbf{Conventional instruction planning methods.} Languages instruction-based policy has been a popular research area in robotics, e.g., visual language navigation, visual question and answer, and daily tasks \cite{XU2021,visionlanguanavigation,alfredshridhar2020}. This paper focuses on long-term daily tasks that involve navigation and interaction. Hierarchical planning methods, which utilize rules or expert guidance to generate plans and employ low-level policies to execute them, have demonstrated their effectiveness \cite{Filmmin2021,prompterinoue2022,llm-plannersong2022}. These methods assume that the agent has domain knowledge or expert assistance in a closed universe. Planning issues are often described using PDDL or answer set programming in recent work \cite{Brewka_Eiter_Truszczyński_2011}. A rule-based, search-based, or sampling-based planning algorithm has worked for mobile robots \cite{Segovia-Aguas_Jiménez_Jonsson_2021,DBLP:conf/iros/DingZZZ20} and robotic manipulators \cite{Cohen_Chitta_Likhachev_2010,DBLP:conf/aips/GarrettLK20,li2024stabilizing, liu2024enhancing}. 
However, these methods rely on specialized symbolic and logical representations of planning, which have limited generalizability and cannot handle unforeseen scenarios.

\textbf{Instruction planning method based on LLM.} Scholars are exploring the use of large language model-based planners as planning systems to tackle limitations in generalization \cite{DBLP:conf/iclr/ZengAICWWTPRSLV23,progpromptsingh2023,ding2023task}. They employ prompt engineering methods to generate robot-friendly subgoals \cite{mmreactyang2023mm,vemprala2023chatgpt}. Some planning methods utilize a procedural language for LLMs \cite{CAPLiangHXXHIFZ23}, while the planning process is conducted in an open-loop manner without access to world information \cite{huang2022language,instruct2acthuang2023,CAPLiangHXXHIFZ23}. Saycan \cite{saycanahn2022can}and Text2Motion \cite{saycanahn2022can,lin2023text2motion} use LLM to predict subgoals and select feasible actions based on environment or geometric constraints. LLMs can dynamically update the robot's plan online by calling LLMs multiple times \cite{llm-plannersong2022}. However, the findings suggest that LLMs performance in long-term task planning is frustrating, limited by feasibility and correction, even in seemingly uncomplicated tasks \cite{saycanahn2022can}. This paper enhances LLM-based planning by incorporating a new robotic dataset and re-planning to boost feasibility and planning accuracy.

\begin{figure*}[htb!]
   \centerline{
         \includegraphics[width=0.95\linewidth]{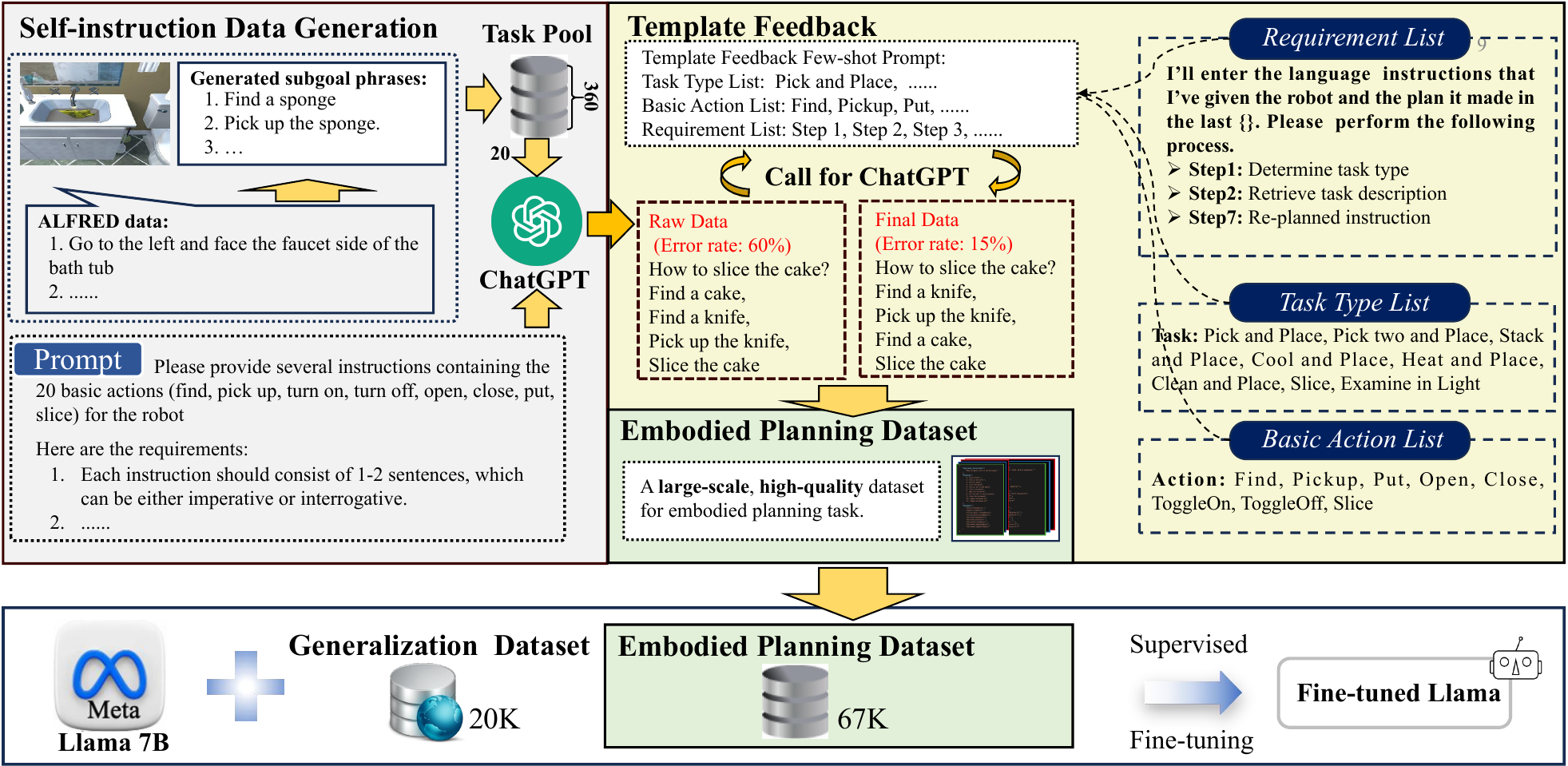}
   }
    \captionsetup{font=small}
\caption{The framework of template feedback-based self-instruction data generation and RoboPlanner training process.
   }
   \label{Fig:datagenerate}
\end{figure*}

\section{RoboGPT System}
A RoboGPT system is proposed for daily instruction following tasks,  consisting of RoboPlanner, RoboSkill, and Re-Plan (see Fig. \ref{fig:architecture}). Given an instruction, RoboPlanner decomposes it into logical subgoals. RoboSkill performs navigation or manipulation skills based on subgoals, produces actions that interact with the environment, and completes all subgoals sequentially.  If a subgoal remains uncompleted, Re-Plan gathers feedback and creates a new plan using the environmental information.
Two key points need to be noted:

\begin{itemize}
\setlength{\itemsep}{0pt}
\setlength{\parsep}{0pt}
\setlength{\parskip}{0pt}
\item Beyond the mundane and repetitive daily tasks, RoboGPT builds a more complex and diverse embodied planning dataset. RoboPlanner, equipped with common-sense knowledge of daily tasks and expertise in robotics, is trained to tackle complex and long-term decision challenges, including the ability to locate concealed objects within containers.

\item Due to the nomenclature diversity between the instruction and the environment, the same object can be referred to using different linguistic representations. Re-Plan addresses the challenge of nomenclature diversity by mapping multiple instruction target representations to corresponding objects in the environment. Additionally, an enhanced perception model is designed to collaborate with Re-Plan, ensuring accurate and feasible re-planning capabilities.
\end{itemize}

\subsection{RoboPlanner}
Although LLMs already demonstrate remarkable performance in general-purpose scenarios, they still underperform in vertical domains like robotics due to a lack of domain expertise and data. Within an academic budget, training an LLM in a specialized domain requires a strong pre-trained LLM and high-quality training data.

Therefore, we build a 67K high-quality embodied planning dataset with template-based self-instruction data generation.  The dataset is used to train a planning model, based on Llama \cite{touvron2023llama}, capable of handling daily tasks with long-term decision making. The process of dataset generation and model training is illustrated in Fig. \ref{Fig:datagenerate}.

Publicly available data for planning robots with long-term decision making in daily tasks is limited. The most relevant ALFRED task \cite{alfredshridhar2020} contains only 8K expert trajectories covering seven task types, which is insufficiently diverse. As a result, we not only derive 60K samples of various types from the ALFRED tasks, but we also employ self-instruction to produce 7K samples with a wider range of task descriptions and types.

\subsubsection{Embodied Planning Dataset} By transforming and expanding the ALFRED task into robot-friendly subgoal phrases, we ultimately obtain an embodied planning dataset of 67K instruction instances. In the ALFRED task, 8K expert trajectories contain high-level instructions (task instructions) and low-level instructions (planning descriptions), e.g., a high-level instruction is: `\textit{Put a clean sponge on a metal rack}', and the low-level instructions are: `\textit{Go to the left and face the faucet side of the bath tub. Pick up the left-most green sponge from the bath tub. Turn around ... left of the lotion bottle}'. The low-level instructions contain a substantial amount of scenario-specific environmental information, which makes it challenging to apply them to different scenarios. Therefore, we convert the detailed, low-level instructions into subgoal phrases. Through analysis, we identify the consistent planning steps for each task and abstract them into a single series of subgoals. For example, rewrite the above low-level instructions as `\textit{find a sponge; pick up the sponge; find a sink; put the sponge in the sink;... put the sponge on the metal rack}'. The planned subgoals are all combinations of robot skills like navigation, grasping, opening, putting, and so on. Therefore, it's adaptable to any robot and task context. 
Based on the seven types of ALFRED tasks, we derive five new tasks and construct planning templates to enrich embodied planning data, generating a total of 60K samples.

The data from ALFRED demonstrations has limitations in terms of diversity (only seven $+$ five types) and quality, which hinders the generality of the tuned model. The suboptimal quality of the data stems from the employment of crowdsourcing techniques for task instructions, which involves three individuals watching videos and subsequently providing their individual descriptions. There are numerous inconsistencies in the descriptions of objects and actions depicted in the videos among the three individuals. Consequently, approximately 20\% of the data descriptions are incorrect \cite{alfredshridhar2020}. While the quantity, diversity, and creativity of human-written instruction data are limited, we tackle this issue by employing a three-step data generation process, resulting in the generation of 7K generalized data.

Step 1: Self-instruction Data Generation. Following Self-Instruct\cite{DBLP:conf/acl/WangKMLSKH23}, we employ ChatGPT to generate and expand the ALFRED dataset in a few-shot setting. We create precise prompts that enable ChatGPT to convert instructions into robot-friendly subgoals. In these prompts, the agent should fulfill the common requirements of various real-world applications, ensuring that the generated steps are practical for a robot to execute. Here's an overview of how ChatGPT generates the data:
Initially, we manually select 360 samples from ALFRED's dataset and add them to the task pool. Subsequently, we randomly select 20 samples from the task pool. The few-shot prompt includes these samples, as well as the custom prompt. Finally, ChatGPT processes the few-shot prompt and generates diverse samples. 

Step 2: Template Feedback. We introduce a template feedback mechanism that prompts ChatGPT to introspect and make corrections to the generated data. Due to the scarcity of embodied instruction planning datasets for generic LLMs, over 60\% of the data generated in step 1 contain issues, primarily incorrect logical relationships in the planning. We integrate the reflective mechanism into the data generation process. Specifically, we utilize template-based feedback to input raw data, basic action list, task type list, and task requirement list into ChatGPT. ChatGPT then engages in reflective analysis to evaluate the quality of the data and categorize it. Finally, we employ template-guided corrective measures to iteratively refine the generated data. The specific details of the template feedback are depicted in Fig. \ref{Fig:datagenerate}.

 Step 3, after the corrections in step 2, only 15\% of the data exhibited issues. We manually rectify these instances, resulting in a final large-scale and high-quality embodied planning dataset comprising a total of 67K samples. 

\subsubsection{Supervised Fine-tuning}
To enhance the embodied instruction planning capability, we employ supervised fine-tuning on Llama \cite{touvron2023llama}. In addition to 67K embodied planning dataset, we augment the robot's training set with 20K generalization dataset obtained from the Internet and train only two episodes, enabling RoboPlanner to learn the robot's planning capabilities and maintain strong generalization skills. Following \cite{taori2023alpaca}, we utilize cross-entropy loss to guide the model's convergence on the embodied tasks.

\begin{equation}
\begin{aligned}
    loss=-\textstyle\sum_{h=1}^{H}y_h\log_{}{\hat{y}_h}
\end{aligned}
\end{equation}

\noindent where $H$ represents the number of classes, $y_h$ denotes the one-hot encoding of the true class of the $h_{th}$ token, and $\hat{y}_h$ is the probability predicted by the model for the $h_{th}$ token belonging to class $h$.

\subsubsection{RoboPlanner} 
After training, RoboPlanner becomes capable of planning both basic and logically challenging tasks.
For instance, given the task $Inst.$ `\textit{Put a pencil and a book on the table}', RoboPlanner first generates $N$ subgoal-instructions $S_n$, which are a series of short phrases, such as `\textit{pick up the pencil}'. Subsequently, RoboPlanner matches $S_n$ into robot-friendly initial binaries $B_n$, for example, `\textit{Pickup Pencil}'. Finally, $B_n$ is transmitted to the RoboSkill and executed sequentially. 

\begin{equation}
\begin{aligned}
    \{S_n\}_{Inst.}^{N}=\textnormal{RoboPlanner}(Inst.)&, B_n ={\textnormal{Match}}(S_n), \\
    &\text{s.t.} \quad n \in \{1, 2, 3,...N\} 
\end{aligned}
\end{equation}
\noindent Moreover, RoboPlanner can also accurately plan and manipulate hidden objects within containers, e.g., `Get a towel out of the cabinet to soak it, then put it on the toilet', as shown in Fig. \ref{fig_visiableobject}.


\begin{figure*}[!ht]
   \centering
   \centerline{
      \includegraphics[width=0.9\linewidth]{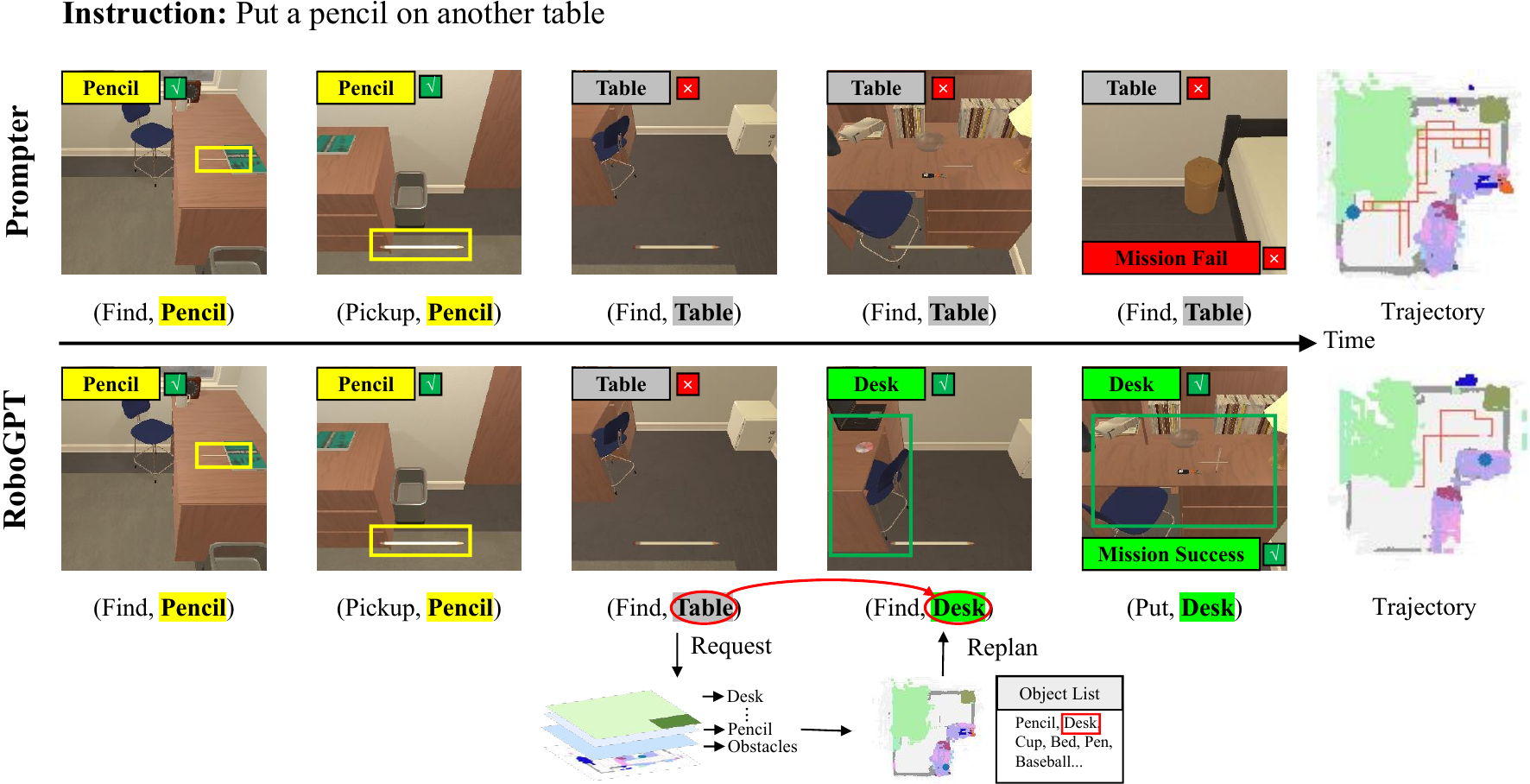}
   }
    \captionsetup{font=small}
   \caption{Effectiveness of Re-Plan. Re-Plan will find similar alternative objects if the subgoal object cannot be noun-aligned with the object already present in the environment.  
   }
   \label{fig:replan}
\end{figure*}

\subsection{RoboSkill}
\label{RoboSkill}
RoboSkill completes navigation or operation skills based on the received instructions. Referring to Prompter \cite{prompterinoue2022}, at each time step, the perception block receives an egocentric RGB image, detects object segmentation and depth, and updates the semantic map of corresponding regions. The semantic map guides the navigation block to look for the target. Once the target is found, the interaction block produces actions to interact with the environment until all subgoals are completed.

Semantic segmentation plays a crucial role in various applications due to its ability to provide object masks for interaction and facilitate the creation of semantic map\cite{semantic02, lu2020cnn,chen2020boost}. The semantic map is subsequently utilized for subgoals such as Re-Plan and object navigation. The segmentation blocks in previous work have been found to exhibit omissions or misidentifications, which have a substantial impact on the accuracy of the semantic map.  Consequently, this leads to a decrease in the performance of Re-Plan and the overall SR. We gather a dataset from ALFRED \cite{alfredshridhar2020} seen environments to train a semantic segmentation model utilizing the FastSAM \cite{zhao2023fast} backbone. This significantly enhances the accuracy of both object detection and the semantic map.

Besides employing more precise models, we also dynamically update the semantic map, which can eliminate the influence of object detection errors to some extent. The global semantic map ${M}^S_t \in [0,1]^{(2+C) \times M \times M}$ is a binary grid map, where $C$ indicates the number of objects and $M \times M$ denotes the number of grids and each grid represents a  5 cm $\times$ 5 cm ground space. We make a dynamic selection of $C$ and include all large objects in the semantic map. However, for small objects, we only consider those that are relevant to the goal and those that can be utilized for Re-Plan purposes. For instance, if the goal is to `\textit{Place a bottle on the desk}' and the subgoal is to `\textit{Find a glass bottle}', we would include objects such as `\textit{glass bottle}', `\textit{wine bottle}', `\textit{soap bottle}', and other types of bottles in the semantic map. 

We employ \textnormal{U-Net} \cite{Unet} and \textnormal{FastSAM} \cite{zhao2023fast} separately to estimate depth $I_t^D$ and segmentation $I_t^S$, and utilize them to generate local semantic map $\hat{M}_t^S$ and binary observation map $M_t^O$: 

\begin{equation}
\label{ex2}
(\hat{M}_{t}^{S}, M_{t}^{O}) = \textnormal{MapModel}\left ( I_{t}^{D},I_{t}^{S} \right)
\end{equation}

\begin{equation}
\label{ex3}
\begin{aligned}
M_{t}^{S} &= \textnormal{MapUpdate}(\hat{M}_{t}^{S},M_{t}^{O}) \\
&= \hat{M}_{t}^{S}\times M_{t}^{O} \times {M}_{nbr} + {M}_{t-1}^{S}\times  \left( 1 - {M}_{nbr} \right) \nonumber
\end{aligned}
\end{equation}
\vspace{-0.6em}
\begin{equation}
\begin{aligned}
\qquad if \  {M}_{nbr}\cap M_t^{O}\neq \varnothing \  and \ {v}_{i} \ not \  in \  M_{nbr}
\end{aligned}
\end{equation}
\vspace{-0.6em}

\noindent Then we update $M_t^S$ using $\hat{M}_t^S$ and $M_t^O$, and finally, we obtain the inventory vector  $\textnormal{V}_t=\textnormal{Object-Detector}(M_t^S)$. Updating ${M}^S_t$ can eliminate the effects of object detection errors, but this correction is not timely. It will only update the neighborhood region $M_{nbr}$ around the target object $v_{i}$ when the agent does not detect $v_{i}$ in the current observation area. This means that if re-planning occurs before the update of $M^S_t$, Re-Plan will use an incorrect semantic map to modify the subgoals. Therefore, to address this particular scenario, RoboGPT employs the number of pixels ${n_{v_i}}$ as a criterion to determine the reliability of confirming the presence of an object in the current scene and gets the confirmed inventory vectors $\bar{\textnormal{V}}_{t}$ from $\textnormal{V}_t$.

\begin{equation}
{\bar{\textnormal{V}}}_t = {[v_i^{'} \  for \  v_i \  in \  {\textnormal{V}}_t \  if \  n_{v_i}>p ]}\
\end{equation}

\subsection{Re-Plan}
One of the primary challenges in embodied AI is nomenclature diversity, where the targets mentioned in instructions and the objects present in the environment may have different names for referring to the same object. However, existing approaches \cite{prompterinoue2022} often neglect this particular issue. They primarily focus on task planning prior to execution, without fully comprehending the current environment. When there is a mismatch between the objects specified in the subgoal and the objects present in the environment, the agent will endlessly explore until it exceeds the maximum number of steps.

LLM-based planners \cite{llm-plannersong2022,progpromptsingh2023} usually input a list of all detected objects into the LLM and provide a new plan, which relies heavily on consuming LLM and image detection, resulting in wasted resources and incorrect re-planning results once detection fails. 


Different from other planners, when RoboGPT fails to find the target even after thorough exploration, it utilizes semantic map for re-planning. Specifically, Re-Plan employs BERT \cite{devlin2018bert} to calculate the similarity between the current target $v_i$ and the set of $\bar{\textnormal{V}}_{t}$.

\begin{equation}
\setlength{\abovedisplayskip}{1pt}
\setlength{\belowdisplayskip}{1pt}
\label{ex4}
{\textnormal{Sim}}\left \{s_1,s_2,....,s_k  \right \}= \textnormal{BERT}\left ( v_i,\bar{\textnormal{V}}_t=\left \{v_1^{'},v_2^{'},....,v_k^{'}\right \}\right )
\end{equation}

The most similar target ($v_j^{'}$,  $j=$ $\textnormal{argmax}({\textnormal{Sim}})$) that satisfies the condition $s_j> 0.7$, is considered to be the same object as $v_i$. Replace $v_i$ in $S_n$ with $v_j^{'}$ to form a new subgoal-instruction $S^{'}_n$. The robot then proceeds to perform this new task. A more intuitive example of re-planning is illustrated in Fig. \ref{fig:replan}. Re-Plan will find similar alternative objects `\textit{Desk}' if the subgoal object `\textit{Table}' cannot be found in the environment.

\section{Experiments}
We conduct a series of experiments to validate the ability of RoboGPT to handle daily tasks on both the ALFRED tasks and the constructed generic tasks. We also investigate and analyze the roles of different modules.
There are several aspects of the proposed RoboGPT that require verification: 1) the static planning capability of RoboPlanner (see Sec.\ref{result_roboplanner}). Verify whether RoboPlanner can effectively plan a series of subgoals to accomplish complex, long-sequence tasks without interacting with the environment; 2) the effectiveness of RoboSkill (see Sec.\ref{result_roboskill}); and 3) Re-Plan capability (see Sec.\ref{resultRe-plane}). During the task, verify whether RoboGPT can re-plan the target objects based on the physical scene to address the issue of open vocabulary; 4) the performance of RoboGPT in solving complex daily tasks (see Sec.\ref{result_alfred} and \ref{reusltAlfredtest});
Hence, we perform a series of experiments and explain experimental setup, and experimental results.

\subsection{Experiemental Setup}
\textbf{Metrics:} 
Following \cite{prompterinoue2022,llm-plannersong2022}, we report the following metrics to assess the effectiveness of RoboGPT:
success rate (SR), goal-condition success (GC), and high-level planning accuracy (HLP ACC). SR is the agent's overall task completion rate. GC is the ratio of completed goal-conditions, e.g., in `\textit{Heating a cleaned apple}', `\textit{washing}' and `\textit{heating}' are goal-conditions. Using SR and GC, the path length weighted SR (PLWSR) and path length weighted GC (PLWGC) are defined as (path length of the expert trajectory)/ (path length taken by the agent). HLP ACC is the accuracy of subgoal planning\cite{llm-plannersong2022}.

\textbf{Baselines:} 
We primarily compare with three kinds of baseline methods. The first is the SOTA algorithm, CAPEAM\cite{CAPEAM2023}, which utilizes the context of task instructions to predict the agent's next subgoal. Another one is Prompter \cite{prompterinoue2022}, which relies on template-based methods for instruction planning. The last is LLM-Planner \cite{llm-plannersong2022} and OPEx \cite{shi2024opex}, which utilize ChatGPT for instruction planning.

First, we reproduce Promper \cite{prompterinoue2022} and LLM-Planner \cite{llm-plannersong2022} to conduct testing on the Valid Unseen tasks with ground truth information. Prompter consists of three components: high-level planner, perceptual mapping, and navigation/interaction policy. The high-level planner used by Prompter is similar to the one employed by FILM \cite{Filmmin2021}. It employs a classifier based on BERT to categorize each task into one of seven task types. Additionally, it predicts parameters associated with each task, including target objects, receptacles, parent objects, and slicing. One notable difference between Prompter and FILM lies in their semantic search modules. Prompter searches target objects based on landmarks, which speeds up the search. In the comparative experiments, the pre-generated subgoals provided by Prompter are utilized without any modifications. LLM-Planner \cite{llm-plannersong2022} consists of two main components, the high-level planner and the re-plan module. These components are integrated into the overall framework. Thus, in the comparative experiments, the high-level planner based on Prompter's template is substituted with LLM-Planner. A total of 140 cases, 20 for each of the seven ALFRED tasks, are chosen. To assess the similarity between each training example and the current test example, the pre-trained BERT-base-uncased model is employed. The concept of similarity is established by calculating the Euclidean distance between the BERT embedding. The k-nearest-neighbor (KNN) retriever is employed to select the nine most comparable in-context cases. These examples are then fed into the LLM to generate a static version of LLM-Planner. To minimize the frequency of LLM calls, we provide LLM-Planner with an inequitable advantage by supplying it with all scene objects prior. Finally, to ensure a fair comparison between each method, we compare their results on the third-party platform ALFRED (Tests Seen and Tests Unseen environments).

\textbf{Evaluation Dataset:} 
The test set has four parts: `Tests Seen' (1533 episodes), `Tests Unseen' (1529 episodes), and `Valid Unseen' (50 episodes) in ALFRED, as well as our own generated `Generalization Task' (50 episodes). Although ALFRED has ground truth for each valid task, we find some of them have issues. Therefore, we select 50 tasks from `Valid Unseen' where the instructions and ground truth are perfectly matched.  The `Generalization Task' consists of 50 complex high-level instructions that we annotate ourselves, and these tasks are entirely distinct from the seven types defined by ALFRED.

\begin{table*}[ht]
\begin{center}
\caption{Performance comparison on seven types of ALFRED tasks (Valid Unseen) and Generalization tasks}
\scalebox{1}{
\begin{tabular}{lcccccc}
\hline
\multirow{2}{*}{\textbf{Method}} & \multicolumn{5}{c}{\textbf{Valid Unseen}} & \textbf{Generalization Task} \\ \cline{2-7} 
                                    & SR                    & PLWSR & GC & PLWGC & HLP ACC  & HLP ACC                      \\ \hline
\multicolumn{5}{l}{High-level Instruction Only}                                                          \\ \hline
Prompter \cite{prompterinoue2022}  & 50 & 20.41 &56.67 &21.86  & 82           & 0       \\

LLM-Planner \cite{llm-plannersong2022}   & 32   &13.54   &47.33  &18.74   & 66        & 34                         \\

\textbf{RoboGPT}                & \textbf{60}  &19.10     & \textbf{69.83}  &21.26  &\textbf{96}     & \textbf{78}   
\\ \hline
    \textbf{RoboGPT w.o. RoboPlanner  }    & 58  & \textbf{22.18} & 65.83 &\textbf{24.17} & 58 &  0  \\
\textbf{RoboGPT w.o. RoboSkill }    & 56 & 21.80 & 64.00 &23.51 & {\textbf{96}} & \textbf{78}                      \\
\textbf{RoboGPT w.o. Re-Plan}      & 52  &16.02   & 62.33 &19.01 & 88   & 78  
\\ \hline
\end{tabular}
}

\label{ablation_test}
\end{center}
\end{table*}

\textbf{Training Details:} 
RoboPlanner is fine-tuned from Llama-7b on NVIDIA DGX A100. Training data includes 67K embodied planning dataset and 20K online generalization dataset. To maintain generalization, the network is trained in 2 episodes with a $10^{-5}$ learning rate. RoboSkill trains a semantic segmentation model leveraging the FastSAM backbone \cite{zhao2023fast} on collected 80K images for 100 epochs with a learning rate of $10^{-3}$, batch size is 16.

\subsection{Data Generation Results}
We expand the ALFRED dataset to 67K using ChatGPT through self-instruction. However, due to the lack of robot instruction planning dataset for ChatGPT, over 60\% of the instructions directly generated by ChatGPT contain logical errors. By implementing template feedback, the proportion of logical errors decreases to 15\%, significantly improving the dataset's quality and reducing the manual calibration workload in later stages. Our method (self-instruction + template feedback) provides a way to generate robot instruction planning data on a large scale.

\subsection{Ablation Experimental Result}

\subsubsection{Effectiveness of RoboPlanner} 
\label{result_roboplanner}
In the third-to-last column of Table \ref{ablation_test}, RoboPlanner is replaced with the Prompter's planner \cite{prompterinoue2022}, resulting in an 8\% decrease in SR and a large decrease in HLP ACC. This indicates that RoboPlanner plays a crucial role in understanding tasks that involve long-term decisions, surpassing the capabilities of Prompter in this aspect.

In the designed Generalization task RoboPlanner performs well in complicated long-term tasks, as shown by its HLP ACC in Table \ref{ablation_test} being far ahead of others. Prompter performs effectively on ALFRED tasks due to its specific template designed for ALFRED, but it lacks generalization to other tasks. On the other hand, LLM-Planner utilizes ChatGPT for logical reasoning, but its performance relies on pre-labeled data for the prompt. Although it demonstrates a certain level of generalization, the performance of LLM-Planner remains unstable.

Prompter and FILM \cite{prompterinoue2022, Filmmin2021} view task planning as a classification task and utilize templates for planning. They may misjudge the task type, the parent target, and the target object when planning a task, e.g., the task is `\textit{Put two potatoes in microwave}', while Prompter plans `\textit{pick up an egg}'. It could be due to overfitting the training data. 

LLM-Planner \cite{llm-plannersong2022} uses ChatGPT for planning and may make logical errors when planning. For instance, when given a task like `\textit{Place a glass with a knife in a sink.}', it plans `\textit{Pick up glass then pick up the knife, and put knife in the sink}', which doesn't put the knife in the glass, failing to understand the relationship between object and container.

\begin{figure*}[!ht]
  \centering
   \centerline{
      \includegraphics[width=0.9\linewidth]{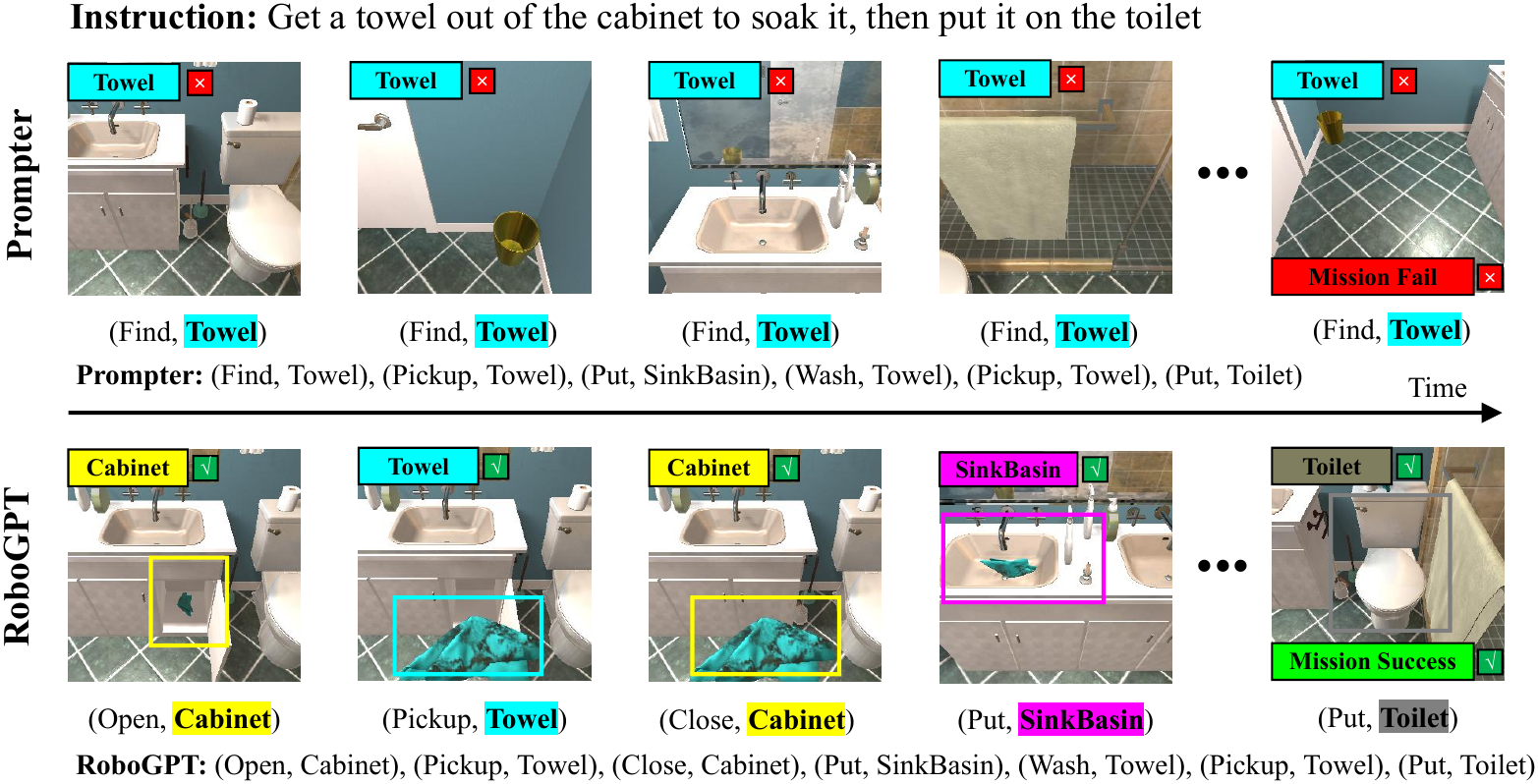}
   }
   \caption{Planning of the task with the invisible object in containers. RoboPlanner possesses the capability to comprehend object relationships, hence facilitating the accomplishment of tasks involving the presence of a target object within an enclosed area.
   }
   \label{fig:planning}
   \label{fig_visiableobject}
\end{figure*}

Beyond the above methods, the advantages of RoboPlanner can be summarized as follows: 
1) \textbf{prefix understood}: using environmental information as a prefix prompt, it can produce practical planning; e.g., for the task `\textit{There is a stove and no microwave, how to heat an apple}', RoboPlanner plans `\textit{find a stove}' to heat the apple, while other planners usually plan to `\textit{find a microwave}';
 2) \textbf{quantity understood}, RoboPlanner comprehends the need for robots to pick up objects individually rather than all at once, and it can plan tasks with 3 or more objects, while other approaches can only plan tasks with 2 objects. e.g., `\textit{Put four books on the desk}'; 
 3) \textbf{object dependencies understood}, RoboGPT can use task instructions to infer the location of invisible objects. For instance, if the task is `\textit{Get a towel out of the cabinet to soak it, then put it on the toilet.}', RoboPlanner  will plan to `\textit{open the cabinet}' first, whereas others will simply find a towel (shown in Fig. \ref{fig_visiableobject});
 4) \textbf{tasks with ultra-long-term decision understood} RoboPlanner can understand and plan complex tasks with more than 30 subgoals. e.g., `\textit{cut a slice of bread, warm it with the microwave, put it on the counter along with putting the knife in the cabinet}'. 

\subsubsection{ Effectiveness of RoboSkill}
\label{result_roboskill}
When comparing RoboGPT without RoboSkill in the last second row of Table \ref{ablation_test}, RoboGPT produces significantly higher SR (+4\% improvement) and GC (+5.83\% improvement), which indicates a major improvement in perception. As a result, RoboSkill can build an accurate semantic map and couple it with Re-Plan. The segmentation and detection abilities of RoboSkill demonstrate superior performance compared to other approaches.
Furthermore, despite the high values of HLP ACC and GC  achieved by RoboGPT, the SR of execution is relatively lower, suggesting that many tasks fail during the interaction. Consequently, there is room for improvement in the interaction algorithm to enhance the robot's SR during the tasks.

\subsubsection{ Effectiveness of Re-Plan}
\label{resultRe-plane}
We employ Re-Plan with low-computing to achieve quick re-planning. Compared with RoboGPT without Re-Plan in the last row of Table \ref{ablation_test}, RoboGPT achieves higher PLWSR and SR and can complete the task in the ALFRED environment with fewer steps, leading to a higher PLWSR. In cases where the target object mentioned in subgoal cannot be noun-aligned with any existing object in the environment, RoboGPT will find suitable alternative objects as substitutions,
 e.g., replace `\textit{table}' with `\textit{desk}'. This demonstrates how Re-Plan aids in understanding the environment and the alignment of objects and targets within instructions. Table \ref{ablation_test} further highlights the critical role of Re-Plan in achieving an 8\% increase in SR.


\subsection{Experimental Results}
\label{result_alfred}

Table \ref{ablation_test} summarizes the performance of RoboGPT and other methods: Prompter \cite{prompterinoue2022} with template-based planner and LLM-Planner \cite{llm-plannersong2022} with ChatGPT planner. RoboGPT achieves 10.00\% absolute (20.00\% relative) gain in SR on Valid Unseen tasks. 
Our RoboGPT also achieves a definite advantage in HLP ACC: 14.0\% absolute (17.0\% relative) gain on seven types of tasks, significantly exceeding other methods. 

Notably, facing the generalization tasks beyond the seven type tasks, RoboGPT outperforms LLM-Planner \cite{llm-plannersong2022} by {\textbf{44\%}}, and Prompter \cite{prompterinoue2022} by {\textbf{78\%! }} This implies that RoboGPT is extremely generalizable and capable of daily tasks of long-term decision making.

RoboGPT without RoboPlanner has the highest PLWSR and PLWGC than others, even for those using RoboGPT. The reason is that Prompter combines some objects with similar meanings (such as desk lamp and floor lamp, butter knife, and knife) based on rules. Although this can significantly improve the execution efficiency, it is difficult to transfer to complex real scenes. Contrary to Prompter, RoboGPT uses Re-Plan to address the issue of target objects but only after gathering a significant amount of environmental data. Although this strategy is more useful in the actual world, it can also decrease the robot's execution efficiency, resulting in lower scores for PLWSR and PLWGC.

 \begin{figure*}[]
   \centerline{
      \includegraphics[width=1\linewidth]{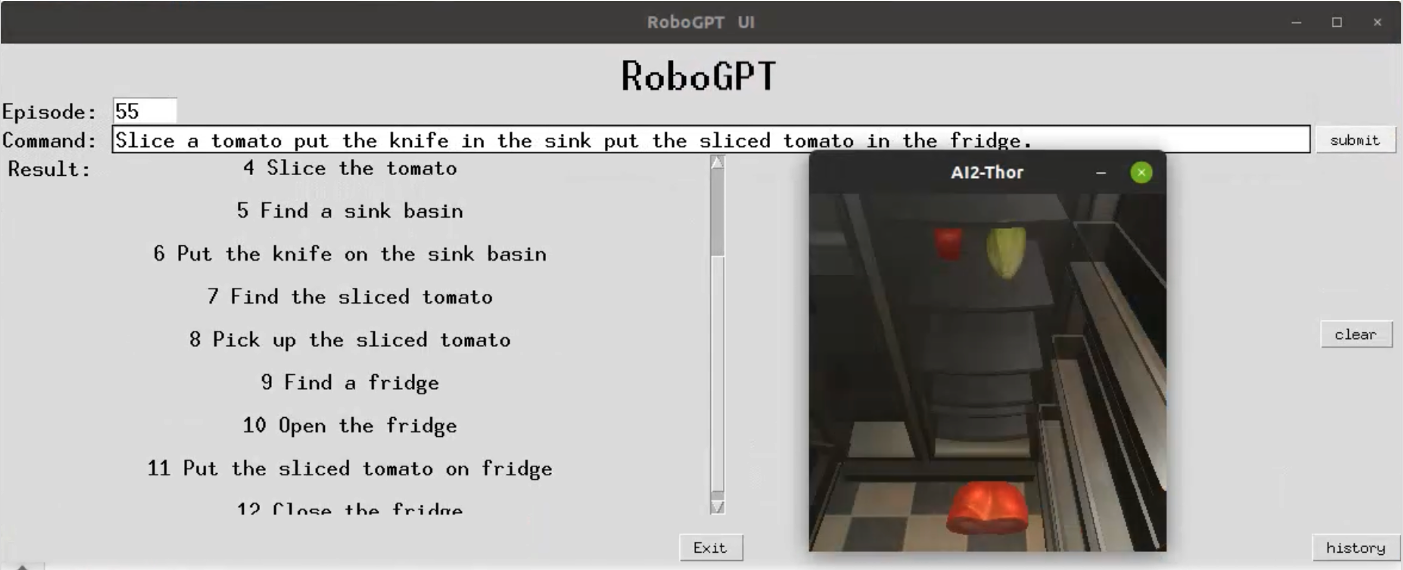}
   }
   \caption{Verification system. The instruction task is `\textit{Slice a tomato, put the knife in the sink, and put the sliced tomato in the fridge}'. The planning of RoboGPT is `\textit{Find a knife, pick up the knife, find a tomato, slice the tomato, find a sink, put the knife in the sink, find the sliced tomato, pick up the sliced tomato, find a fridge, open the fridge, put the sliced tomato in the fridge, close the fridge}'. 
   }
   \label{fig:demo}
 
\end{figure*}

\subsection{Experimental Results on ALFRED test tasks}
\label{reusltAlfredtest}

\begin{table}[]
\begin{center}
\caption{ Performance comparison on ALFRED test tasks. 
}
\begin{tabular}{lllll}
\hline
\multicolumn{1}{c}{\multirow{2}{*}{Methods}} & \multicolumn{2}{l}{Tests Seen}  & \multicolumn{2}{l}{Tests Unseen} \\ \cline{2-5} 
\multicolumn{1}{c}{}                         & GC             & SR             & GC              & SR             \\ \hline
\multicolumn{5}{l}{High+Low-level instructions}                                                                   \\ \hline
HLSM \cite{HLSM2021}                   & 41.21          & 29.94          & 30.31           & 20.27          \\
LGS-RPA \cite{LSG-RPA2022}              & 48.66          & 40.05          & 45.24           & 35.41          \\
ET \cite{ET-PashevichS021}                 & 45.44          & 38.42          & 18.56           & 8.57           \\
MCR-Agent \cite{MCR-agent2023}              & -              & 30.13          & -               & 17.04          \\
M-TRACK \cite{M-TRACK2022}                   & 22.60          & 16.29          & 33.35           & 24.79          \\
LEBP     \cite{LEBP2022}                                    & 36.79          & 28.30          & 36.33           & 28.97          \\
Prompter    \cite{prompterinoue2022}                                 & 60.22          & 51.17          & 56.57           & 45.32          \\
CAPEAM        \cite{CAPEAM2023}                               & 60.98          & 52.58          & 61.40           & 50.36          \\
\textbf{RoboGPT (Ours)    }                     & \textbf{69.24} & \textbf{61.77} & \textbf{70.53}  & \textbf{60.76} \\ \hline
\multicolumn{5}{l}{High level instruction only}                                                                   \\ \hline
EPA \cite{EPA2022}                     & 44.14          & 39.96          & 39.54           & 36.07          \\
HLSM \cite{HLSM2021}                  & 35.79          & 25.11          & 27.24           & 16.29          \\
FILM \cite{Filmmin2021}                     & 36.15          & 25.77          & 34.75           & 24.46          \\
Prompter \cite{prompterinoue2022}             & 56.98          & 47.95          & 53.69           & 41.53          \\
LLM-Planner \cite{llm-plannersong2022}             & 26.77          & 18.20          & 23.37           & 16.42          \\
OPEx \cite{shi2024opex}             & 54.27          & 43.51          & 53.82           & 41.27          \\
\textbf{RoboGPT (Ours)    }          & \textbf{60.25} & \textbf{52.38} & \textbf{65.30}  & \textbf{55.46} \\ \hline
\label{table_test_result}
\end{tabular}
\end{center}
\end{table}

To ensure a fair comparison of performance across all methods, we conduct our testing on the third-party platform ALFRED. We evaluate our approach on the ALFRED test tasks and achieve SOTA results in both seen and unseen environments (shown in Table \ref{table_test_result}).
Compared to methods requiring low-level step-by-step instructions \cite{MCR-agent2023,LSG-RPA2022} and the method requiring ChatGPT planning \cite{llm-plannersong2022}, RoboGPT achieves the best performance with at least 10.4\% improvement of SR in Test Unseen. This demonstrates the planning ability of RoboGPT even beyond the step-by-step guidance of an expert or ChatGPT. In reality, the limitations of the ALFRED dataset hinder RoboGPT from achieving higher SR. ALFRED is a crowdsourced form of labeled data, with many irregularities in the labeling. By our count, more than 20\% of the high-level instructions (instruction tasks) are ambiguous or even incorrect. 
Wrong instruction tasks lead to errors in our planning. This explains why there is a difference in SR of different methods between Table \ref{ablation_test} and Table \ref{table_test_result}. The Prompter initially categorizes tasks and then plans accordingly using templates. However, due to overfitting, it may mistakenly categorize certain erroneous instructions under the correct types, resulting in appropriate planning outcomes. 

We develop a verification system based on the ALFRED  simulation system. You can input arbitrary natural language commands and use RoboGPT to interact with the environment to accomplish the task. The demo is shown in  Fig. \ref{fig:demo}. The instruction task is `\textit{Slice a tomato, put the knife in the sink and put the sliced tomato in the fridge}', and RoboPlanner plans  `\textit{Find a knife, pick up the knife, find a tomato, slice the tomato, find a sink, put the knife in the sink, find the sliced tomato, pick up the sliced tomato, find a fridge, open the fridge, put the sliced tomato in the fridge, close the fridge}'. 

\section{Conclusions}
In this paper, we design a RoboGPT for solving daily instruction following tasks with long-term decisions. The planning module RoboPlanner enhances and fine-tunes Llama using the collected 67K robotic dataset to integrate the world knowledge of LLMs with the expert knowledge of robots, which can understand the prefix context, object quantities, object dependencies, and the tasks with long-term decisions, handling most of daily tasks. Roboskill with an accurate perception model FastSAM is developed, resulting in improved navigation and manipulation abilities. Additionally, the designed Re-Plan adapts the planning to the environment, mitigating the nomenclature diversity problem. The paper provides a well-generalized method for daily instruction following tasks with long-term decisions in robotics. Future improvements may focus on multi-modal embodied planning and manipulation.

\bibliographystyle{IEEEtran}
\bibliography{RoboGPT}

\end{document}